\pdfoutput=1

\documentclass[11pt,table]{article}
\usepackage{siunitx}
\usepackage{acl}
\usepackage{times}
\usepackage{latexsym}
\usepackage{amsmath}
\usepackage{booktabs}
\usepackage{multirow}
\usepackage{xr}
\usepackage{amssymb} 
\definecolor{mycolor}{RGB}{222,232,225} %

\usepackage{graphicx}
\usepackage{subcaption} 

\usepackage[T1]{fontenc}

\usepackage[utf8]{inputenc}

\usepackage{microtype}

\usepackage{inconsolata}

\usepackage{graphicx}

\title{TextVidBench: A Benchmark for Long Video Scene Text Understanding}

\author{
  \begin{tabular}{c}
    Yangyang Zhong\textsuperscript{1} ,
    Ji Qi\textsuperscript{2} ,
    Yuan Yao\textsuperscript{2} ,
    Pengxin Luo\textsuperscript{1} ,
    Yunfeng Yan\textsuperscript{1} \\ 
    Donglian Qi\textsuperscript{1} ,
    Zhiyuan Liu\textsuperscript{2}  ,
    Tat-Seng Chua\textsuperscript{3}
  \end{tabular}
}

\begin{document}
\maketitle
\begin{abstract}
Despite recent progress on the short-video Text-Visual Question Answering (ViteVQA) task—largely driven by benchmarks such as M4-ViteVQA—existing datasets still suffer from limited video duration and narrow evaluation scopes, making it difficult to adequately assess the growing capabilities of powerful multimodal large language models (MLLMs). To address these limitations, we introduce TextVidBench , the first benchmark specifically designed for long-video text question answering ($>$3 minutes). TextVidBench makes three key contributions:1) Cross-domain long-video coverage : Spanning 9 categories (e.g., news, sports, gaming), with an average video length of 2306 seconds, enabling more realistic evaluation of long-video understanding. 2) A three-stage evaluation framework : ``Text Needle-in-Haystack $\rightarrow$ Temporal Grounding $\rightarrow$ Text Dynamics Captioning''.3) High-quality fine-grained annotations : Containing over 5,000 question-answer pairs with detailed semantic labeling.Furthermore, we propose an efficient paradigm for improving large models through: (i) introducing the IT-Rope mechanism and temporal prompt engineering to enhance temporal perception, (ii) adopting non-uniform positional encoding to better handle long video sequences, and (iii) applying lightweight fine-tuning on video-text data. Extensive experiments on multiple public datasets as well as TextVidBench demonstrate that our new benchmark presents significant challenges to existing models, while our proposed method offers valuable insights into improving long-video scene text understanding capabilities.

\end{abstract}

\section{Introduction}
Visual Question Answering (VQA) is often regarded as the "Turing test" for image and video understanding. Early research advanced the methodology by constructing diverse VQA datasets \cite{vqa1,vqa2,vqa3,vqa4,vqa5,vqa10,vqa6,vqa7,vqa8,vqa9}. However, traditional VQA tasks primarily focus on reasoning about objects, scenes, and actions, largely neglecting scene text—a critical component of visual perception. This limitation introduces a significant gap with real-world scenarios: according to statistics from MS-COCO \cite{cocotext}, approximately 50\% of everyday images contain textual elements, while dynamic text in videos—such as road signs, timers, and scoreboards—further enrichs the semantic content.
n recent years, the task of Video Text Visual Question Answering (ViteVQA) has gained increasing attention. M4-ViteVQA\cite{zhao2022towards} was the first to systematically exploit spatio-temporal textual cues in videos for question answering, with benchmarks covering scenarios such as education and driving. This work highlighted the necessity of cross-modal text-visual understanding. Subsequent studies have further explored scene-specific adaptations: NewsVideoQA \cite{2022Watching} focuses on textual information in news videos, while RoadTextVQA~\cite{tom2023readinglanestextvideoqa} targets road signs and street texts in autonomous driving contexts.

However, existing approaches face a key limitation—short-term dependency bias . Current benchmark videos are typically less than 10 seconds long on average (e.g., M4-ViteVQA~\cite{zhao2022towards}), and such short clips rarely involve scene transitions, making it possible to complete tasks by sampling only a few frames. In contrast, real-world applications often require reasoning over long-term temporal contexts (e.g., videos longer than 3 minutes). For instance, in a 30-minute lecture video, models may need to locate a specific few-second segment based on relevant textual knowledge, or track score changes of an athlete across tens of minutes in a sports video. Current benchmarks for video text understanding lack the design to evaluate such long-term temporal reasoning capabilities, which significantly limits the assessment and improvement of models in effectively utilizing textual information from extended video content.

To address the aforementioned challenges, we propose the first benchmark for long-video scene text understanding (TextVidBench) and explore several effective strategies to enhance long-video text comprehension capabilities. \textbf{Our main contributions are as follows:}

\vspace{1mm}

\noindent \textbf{1) TextVidBench.} 
Covering 30 hours of videos from 9 diverse scenarios (including driving, sports, gaming, etc.), with an average video length of 2,306 seconds (100× longer than existing benchmarks~\cite{zhao2022towards}). The benchmark includes three evaluation dimensions (details in Section~\ref{sec:benchmark}):
\begin{itemize}
    \item \textbf{Text Needle-in-Haystack}: Long-video text retrieval (e.g., "In a certain frame, a question appears with the text 'xxx.' What is option 'a' below it?").
    \item \textbf{Text Temporal Grounding}: Text timestamp prediction (e.g., "Find a frame showing two tall buildings with a yellow truck in front, labeled 'Truck.' At what second does this frame appear?").
    \item \textbf{Text Dynamics Captioning}: Dynamic text evolution description (e.g., "How did the game score change over time?").
\end{itemize}

\noindent \textbf{2) Method-Level Explorations.}
We investigate key factors influencing long-video scene text understanding, primarily focusing on fundamental text recognition, temporal awareness, and long-video processing capabilities. Building upon MiniCPM-V 2.6~\cite{yao2024minicpm} (which already excels in short-video text understanding), we propose the following improvements:
\begin{itemize}
   \item \textbf{Integrating Temporal Rotary Position Embedding (IT-RoPE)}: Unlike traditional RoPE, IT-RoPE encodes sequential order for each video frame at the language model level, enhancing performance in temporally related video-text tasks (details in Section~\ref{sec:itrope})..
   \item \textbf{Time Prompt}: Injecting sampling metadata (e.g., "Total video length: 640s, uniformly sampled 64 frames at 0s, 10s, ...") improves the model’s adaptability to videos of varying lengths under dynamic sampling rates (details in Section~\ref{sec:timeprompt}).
   \item \textbf{Non-Uniform Position Interpolation (NUPI)}: Incorporating LongRoPE~\cite{longrope}, we find it simple yet effective in extending the model’s capability to process longer video sequences (details in Section~\ref{sec:nonupi}).
\end{itemize}

\noindent \textbf{3) Experimental Validation.}
We evaluate existing mainstream multimodal large models on the proposed benchmark, providing a quantitative assessment of their video scene text understanding capabilities. The results reveal that long-video text understanding remains highly challenging for current models, with generally low evaluation scores. Additionally, our proposed model-level improvements significantly enhance the base model, offering valuable insights for advancing long-video scene text understanding.

\section{Related Work}

\subsection{VideoQA}
Building upon the foundation of traditional Visual Question Answering (VQA), Video Question Answering (VideoQA) extends the paradigm to address questions pertaining to video content, necessitating models to exhibit robust spatiotemporal reasoning abilities. A variety of datasets~\cite{vqa8,vqa9,17,18,19,20,21,22} have been curated to facilitate this endeavor. Notably, MOVIE-QA~\cite{vqa8} and TVQA~\cite{lei2018tvqa} leverage scenes extracted from movies and television programs, whereas the SUTD-TrafficQA~\cite{xu2021sutd} dataset comprises multiple-choice questions centered around diverse traffic events. These datasets comprise video clips from various scenarios, with all questions focusing solely on the visual content of the videos, without taking into account the rich scene text information present within them.

\subsection{VideoQA involving video text}
NewsVideoQA~\cite{2022Watching}, M4-ViteVQA~\cite{zhao2022towards}, and RoadTextVQA~\cite{tom2023readinglanestextvideoqa} are three datasets that require understanding textual information in videos to answer questions. NewsVideoQA focuses on news videos, where a significant amount of information is conveyed through text, demanding the QA system to integrate both visual and textual cues from the video to respond to related questions. RoadTextVQA, on the other hand, centers on driving scenarios, with all questions based on text or road signs appearing in driving videos, which are often challenging to recognize due to occlusion, blur, and perspective distortion. M4-ViteVQA covers videos from various domains, including shopping, driving, sports, movies, and vlogs.

In terms of model design, M4-ViteVQA integrates visual perception modules such as OCR and Faster RCNN~\cite{rcnn}, combined with a language model. Similarly, NewsVideoQA incorporates an OCR module and specifically designs a loss function for OCR tasks during training. RoadTextVQA employs BERT~\cite{devlin2019bert} as its language model, whose bidirectional Transformer encoder architecture limits its adaptability to general conversational tasks. The architectural designs of these models constrain their applicability to some extent, making it difficult to extend them to more complex video understanding tasks.

\subsection{Multimodal Large Model for Video Understanding}
Recent advances in Multimodal Large Language Models (MLLMs) have significantly advanced video understanding~\cite{yao2024minicpm,xue2024longvila,zhang2024long,li2024llava}. This section focuses on MLLMs with strong textual reasoning and long video processing capabilities, particularly 7B-scale architectures suitable for practical deployment, aligning with our focus on long video-text comprehension.

MiniCPMv2.6 is a versatile multimodal model, excelling in single-frame, multi-frame, and video understanding. Its state-of-the-art performance in scene text recognition (STR) makes it an ideal foundation for our work, which extends and optimizes this model.

For long video understanding, LongVA~\cite{zhang2024long} and LongVila~\cite{xue2024longvila} propose a "Needle-in-a-Haystack" evaluation protocol, handling videos of up to 3,000 frames. However, their approach inserts semantically unrelated image frames, creating artificial challenges that diverge from natural video-text understanding. While demonstrating long-context processing feasibility, this synthetic design limits real-world applicability. To address this, our Text Needle-in-Haystack task eliminates irrelevant frame insertions, directly leveraging original video frames to better reflect real-world video-text alignment needs.

\section{Benchmark}

\label{sec:benchmark}
In this section, we first describe the methodology employed for collecting and annotating the videos. Subsequently, we introduce the three evaluation settings along with their corresponding metrics. Finally, we present the statistical data and analysis results of the evaluation, comparing them with several related datasets to highlight the uniqueness and advancement of the benchmark in the field of long-video text question answering.

\subsection{Data Collection and Annotation}

\textbf{Video Collection}
The video dataset for evaluation encompasses nine domains: driving, egocentric, entertainment, game, knowledge, life record, sports, talk shows, and video news. To construct this evaluation dataset, we acquired video samples rich in textual information from the YouTube platform and meticulously selected a series of representative video clips, totaling approximately 23 hours. Each category was allocated an average of about 2 hours of content.

\vspace{2mm}

\noindent{\textbf{Semi-automatic Annotation.}}
To annotate question-answer pairs related to scene text information from the 23-hour video, this paper proposes a semi-automatic annotation framework, as illustrated in Figure \ref{fig:dataproduce}. The specific implementation steps of the framework are as follows: 1) Frame sampling is performed on the video at a rate of 1 frame per second (fps). 2) The lightweight multimodal large model MiniCPM-V \cite{yao2024minicpm}, which possesses strong text comprehension capabilities, is employed to analyze the extracted frames and determine whether text information exists. Frames containing high-quality text information are retained, while others are filtered out. 3) Highly similar adjacent frames are filtered out. 4) GPT-4o \cite{openai2024gpt4technicalreport} is utilized to act as a question designer, generating text-related questions based on the remaining video frames. 5) Researchers with a background in multimodal large models are invited to verify and revise the generated question-answer pairs, with approximately 20\% of the data being corrected.

\begin{figure}[ht]
\centering
\includegraphics[width=\columnwidth]{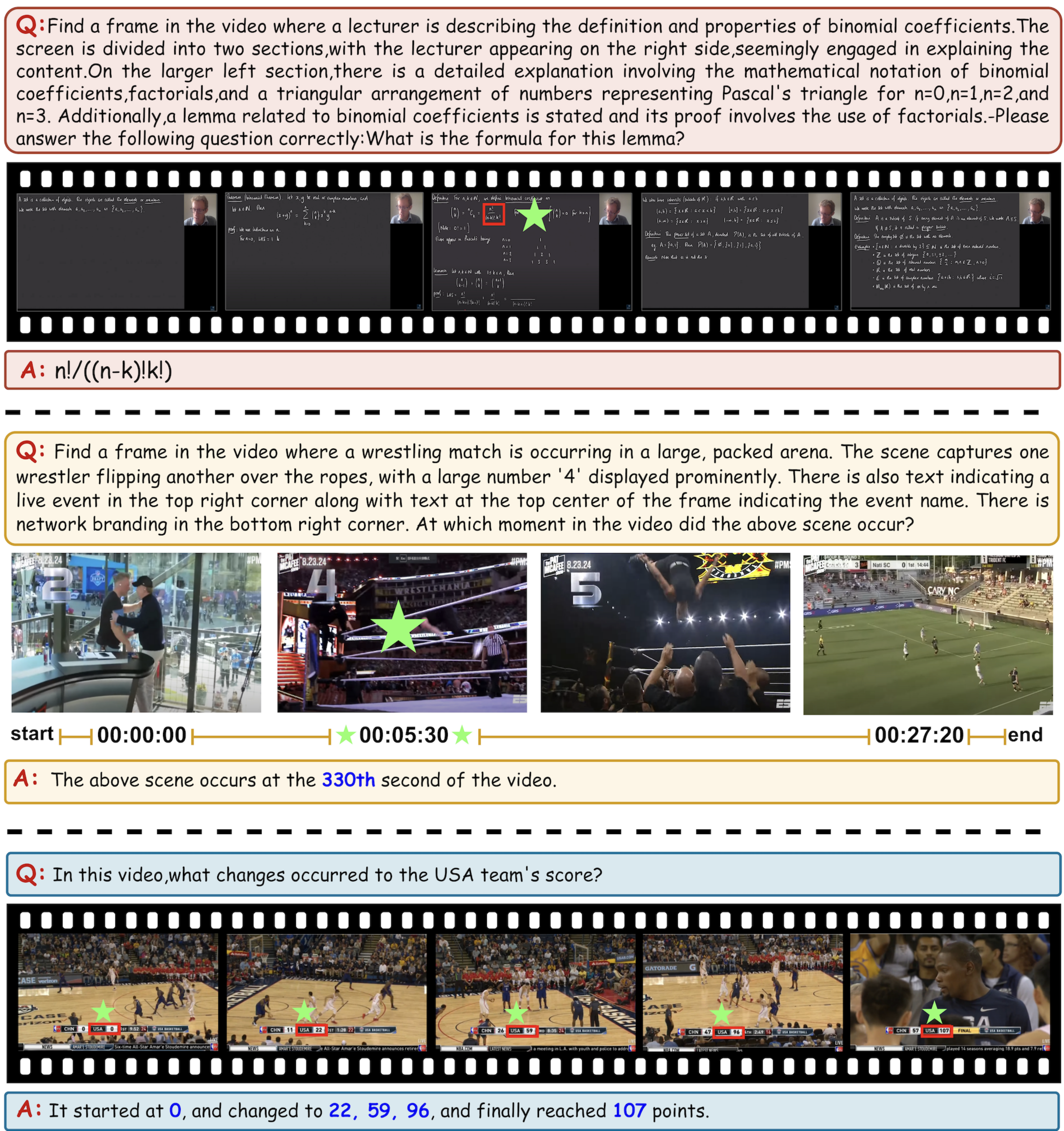} 
\caption{The figure illustrates three distinct task settings, arranged from top to bottom: (1) Text Needle-in-a-Haystack, (2) Temporal Localization, and (3) Text Dynamics Caption.}
\label{fig:bench_case}
\end{figure}

\subsection{Evaluation Setup and Metrics}
\label{sec:Evaluation Setup and Metrics}
This section describes three evaluation setups and their corresponding metrics for assessing video text understanding capabilities.

\noindent{\textbf{1) Text Needle-in-Haystack}}:Existing video-text understanding benchmark datasets \cite{2022Watching,tom2023readinglanestextvideoqa,zhao2022towards} typically utilize video clips with an average duration of 10 seconds. In such short-duration videos, the scene hardly undergoes significant changes, making it almost equivalent to single-image text comprehension. However, video content on actual online platforms often exceeds 3 minutes, rendering existing benchmarks inadequate for effectively evaluating model comprehension in long-video scenarios. To address this, we have constructed a long video-text understanding task with an average duration of 30 minutes, termed the "Text-Needle-in-a-Haystack" task. Unlike tasks such as LongVA \cite{zhang2024long} and Longvila \cite{xue2024longvila}, which involve inserting irrelevant image frames into videos for retrieval, our task design is more aligned with real-world applications: it requires the model to precisely locate the corresponding frame in the original video based on a given text description and answer related text comprehension questions. As illustrated in Fig. \ref{fig:bench_case}, using instructional videos as an example, the model needs to locate specific frames based on course content descriptions and accurately answer questions related to the teaching material. This task setup not only evaluates the model's comprehension ability in long videos with frequent scene changes but also further emphasizes the understanding and application of textual content.

The metric used to evaluate model performance is the ANLS \cite{biten2019scene}, which measures the similarity between the model's output and the ground-truth text.

\noindent{\textbf{2) Text Temporal Grounding:}} In the Text-Needle-in-a-Haystack task, the model is required to accurately identify key frames that match a given textual description from a large number of unordered video frames. However, existing evaluation methods primarily focus on whether the model can locate the correct frame, neglecting the model's ability to identify the specific temporal position of the frame within the sequence, i.e., its temporal localization capability. This capability is crucial in practical applications. For instance, after watching a video, a user might want to revisit and locate a specific moment in the video based on memory. Therefore, we have extended the evaluation framework of the existing Text-Needle-in-a-Haystack task by incorporating a module specifically designed to assess temporal localization capabilities, thereby providing a more comprehensive measure of the model's performance in real-world scenarios. A concrete example is illustrated in Figure \ref{fig:bench_case}.

Evaluation Metrics: Considering that these are long videos (average duration 2306s), we allow a certain margin of error, divided into three levels: 30s/60s/120s. That is, if the timing is within this range compared to the standard answer, it will be considered as correctly located.

\noindent{\textbf{3) Text Dynamics Caption:}} The textual change information in videos holds significant practical value. Taking sports videos as an example, the real-time recognition of match scores is crucial for video commentary. Based on this, we have designed an evaluation task specifically aimed at understanding textual changes in videos. We have selected two types of scenes closely related to textual changes: gaming and sports. The gaming scene is further divided into three specific games: PUBG, League of Legends (LoL), and Dota; the sports scene is subdivided into swimming, basketball, and table tennis. This evaluation task aims to assess the completeness of the model's description of changes in match scores and in-game currency. 

The evaluation method employs GPT-4 as a scoring tool, comparing the model's output with standard answers and assigning a score from 0 to 10 based on the thoroughness of the description.

\subsection{Statistics and Analysis}
We presents a systematic analysis of the statistical characteristics of the constructed dataset. Figure \ref{fig:short}illustrates the analysis across multiple dimensions; please refer to the Appendix section \ref{fulu:Statistics and Analysis} for more details.

\section{Method}

\subsection{Model Architecture}
The architecture of the model we designed is illustrated in Figure \ref{arch}. The detailed description of the structure can be found in the Appendix \ref{fulu:Model Architecture}.


    
    


\subsection{Video Instruction Data Generation and Tuning}

Due to space limitations, the methodology for generating the video instruction dataset and the tuning approach used in this context are detailed in the Appendix section \ref{instructiongeneration}.

\subsection{Integrating Temporal Rotary Position Embedding}
\label{sec:itrope}



In prior research\cite{qwen,llama2,llama3}, the one-dimensional rotary position encoding (1D-RoPE) \cite{su2024roformer} has been predominantly applied to encode positional information along a single sequence dimension in large language models (LLMs). However, when processing video modalities, flattening visual tokens into a one-dimensional sequence not only fails to effectively capture temporal positional information across video frames but may also significantly increase the overall sequence length, thereby complicating the understanding of long-range temporal content.

While recent state-of-the-art models proposed on arXiv platforms, such as ~\cite{qwen25vl}, have introduced multi-dimensional RoPE variants (e.g., mRoPE) that support temporal encoding, most widely adopted model architectures in practical domains still rely on the traditional 1D-RoPE design. Replacing well-established LLMs with these newly proposed architectures often entails prohibitively high pre-training costs.

Given that this work aims to investigate key factors influencing text understanding capabilities in long video scenarios, we propose a novel temporal-aware rotary position encoding scheme—Inflated Temporal Rotary Position Encoding (IT-RoPE) —by extending the conventional 1D-RoPE with explicit temporal encoding components. This design allows the new positional encoding to be directly substituted into legacy model architectures with minimal impact on their original performance. With only light fine-tuning, the enhanced architecture demonstrates significantly improved temporal perception. Detailed implementation can be found in Appendix \ref{Details of IT-RoPE}.

\subsection{Non-uniform Position Interpolation}
\label{sec:nonupi}
The previous chapter discussed IT-RoPE, which improves the efficiency of the original model architecture in processing video frame sequences. This chapter focuses on how to extend the model’s capability to handle long visual token sequences in a cost-effective manner. We draw inspiration from LongRoPE\cite{longrope}, which employs an evolutionary search algorithm on the LLaMA architecture to directly discover a set of scaling coefficients for the 2D rotation matrices in RoPE. These coefficients significantly enhance the model's performance on longer token sequences. We adapt this algorithm into our proposed IT-RoPE framework and, through a similar search process, obtain a set of parameters well-suited to our architecture. This enables a simple yet effective improvement in handling long video sequences. See Appendix \ref{Details of Non-uniform Position Interpolation} for details.

\subsection{Time Prompt For Video Temporal Location}
\label{sec:timeprompt}
In previous sections, we explored methods to enable LLMs to perceive the positional information of video frame sequences and extend their token processing capabilities. However, for real-world applications involving visual question answering on videos, it is also crucial for the model to align the frame sequence with actual temporal timestamps. In short video scenarios (e.g., <60 seconds), it is feasible to fix the sampling frequency and achieve accurate alignment between frames and timestamps through either model architecture design or training data engineering.

When dealing with longer videos—ranging from several minutes to even an hour—dynamic sampling becomes necessary. This introduces uncertainty in the sampling process, making it impractical to rely solely on fixed datasets for training. Inspired by the system prompt design in LLaVA-OneVision~\cite{li2024llava}, we propose providing the model with explicit temporal context, including the total video duration, the number of sampled frames, and the exact timestamp corresponding to each frame. We argue that this approach naturally accommodates dynamic sampling strategies. The specific prompt format is as follows:

\textit{The video has a total duration of \texttt{<DURATION>} seconds, from which \texttt{<FRAME\_COUNT>} frames were uniformly sampled. The corresponding temporal positions of these frames are: \texttt{<TIMESTAMPS>}. Please answer the following question based on this video.}



\section{Experiments}

\subsection{Implementation Details}
The specific details can be found in the Appendix section \ref{Implementationdetailes}

\subsection{Quantitative evaluation}
In this section, we initially validate the effectiveness of the proposed method on existing benchmarks for short video text comprehension. The experimental results demonstrate that our approach is not only fully compatible with the short video scenario but also achieves a significant performance improvement compared to the baseline models in the benchmark methods. Subsequently, we conduct comprehensive comparative experiments on the newly proposed BenchMark, evaluating the performance of current mainstream multimodal large models as well as our proposed model on this benchmark, followed by an in-depth analysis and discussion.

\subsubsection{Validation on Public Benchmark}
We evaluate the performance of our model using the M4-ViteVQA~\cite{zhao2022towards} benchmark dataset. M4-ViteVQA is the first benchmark specifically designed for the Video Text Visual Question Answering (ViteVQA) task, which aims to answer questions by spatiotemporally reasoning over both textual and visual information in videos. The dataset covers a wide range of real-world scenarios, including nine categories such as shopping, traveling, driving, and advertisement. The test set consists of 3,183 question-answer pairs, demonstrating high diversity and representativeness. To comprehensively assess the model's performance on the ViteVQA task, we employ Accuracy and Average Normalized Levenshtein Similarity (ANLS)~\cite{biten2019scene} as evaluation metrics.

\begin{table}[ht]
\centering
\caption{Performance comparison on the M4-ViteVQA.Among them, "Random" refers to selecting OCRs randomly from the videos, and "Human" refers to answers provided by humans.}
\resizebox{0.4\textwidth}{!}{
\begin{tabular}{l|cc|cc}
\hline
\textbf{\multirow{2}{*}{Model}} & \multicolumn{2}{c|}{Val} & \multicolumn{2}{c}{Test} \\  \cline{2-5}
&ACC &ANLS &ACC &ANLS
\\ \hline
Random &0.56 &0.021 &0.60 &0.025 \\ 
Human &78.08 &0.825 &85.27 &0.893 \\ 
M4C~\cite{m4c} &18.66 &0.242 &17.91 &0.238 \\ 
T5-ViteVQA~\cite{zhao2022towards} &23.17 &0.301 &22.17 &0.291  \\ 
LongVA~\cite{zhang2024long} &13.16 &0.270 &12.59 &0.264  \\ 
LongVILA~\cite{xue2024longvila} &18.98 &0.343 &18.53 &0.322  \\ 
MiniCPM-V 2.6 &{32.1} &{0.505} &{32.2} &{0.510}  \\
\rowcolor{mycolor}
Ours &\textbf{38.2} &\textbf{0.561} &\textbf{37.6} &\textbf{0.553} \\ 

\hline
\end{tabular}
}
\label{tab:vtvqabench}
\end{table}

As shown in Table~\ref{tab:vtvqabench}, the proposed model demonstrates significantly superior performance on existing public video-text understanding datasets compared to state-of-the-art video-text understanding expert models (including M4C and T5-ViteVQA) and the latest large language model for long video understanding, LongVA, achieving remarkable improvements in both accuracy and Average Normalized Levenshtein Similarity (ANLS) metrics.

\subsubsection{Validation on Proposed Benchmark}
Section~\ref{sec:benchmark} has provided a comprehensive overview of the datasets and evaluation protocols for benchmarking models in the field of video-text understanding. In this subsection, we detail the proposed model and compare it with other state-of-the-art models, presenting experimental results under the three configurations described in Section~\ref{sec:Evaluation Setup and Metrics}.

\vspace{1mm}

\noindent{\textbf{1) Text Needle-in-Haystack:}}
To thoroughly investigate the model's text comprehension capabilities across varying video durations and diverse scenarios, we designed two sets of experimental tables. As shown in Table \ref{tab:laozhen_time}. As can be seen, our method outperforms other state-of-the-art models; for a detailed analysis, please refer to the Appendix section\ref{Validation on Proposed Benchmark}.

\noindent{\textbf{2) Text Temporal Grounding:}} As shown in Table~\ref{tab:timeloc}, our method demonstrates superior performance in the Temporal Grounding task compared to other state-of-the-art models. For a detailed analysis, please refer to the Appendix section~\ref{Validation on Proposed Benchmark}.

\noindent{\textbf{3) Text Dynamics Captioning:}}
Table \ref{tab:textchange} presents the experimental results on the  Text Dynamics Captioning. A detailed analysis of these experimental results is provided in Appendix section~\ref{Validation on Proposed Benchmark}.

\begin{table*}[h]
    \centering
    \caption{Comparative Evaluation of Model Performance on Video Text Question Answering Across Varied Temporal Lengths \textbf{(Text Needle-in-Haystack Task, fps=0.5)}. Accuracy is quantified in percentage terms (e.g., 9.7 corresponds to 9.7\%).}
    \label{tab:laozhen_time}
    \resizebox{0.8\textwidth}{!}{
    \begin{tabular}{l|ccccccccccc}
        \hline
        \textbf{Models}  & {\textbf{2min}}& {\textbf{4min}}& {\textbf{6min}}&{\textbf{8min}}&{\textbf{10min}}&{\textbf{12min}}&{\textbf{14min}}&{\textbf{16min}}&{\textbf{18min}}&{\textbf{20min}}&{\textbf{Avg}}
        \\
        \hline
        LongVA & 0.263 & 0.251 & 0.245 & 0.244 & 0.244 & 0.245 & 0.238 & 0.230& 0.224 & 0.182 &0.240  \\
        LongVILA-8B & 0.310 & 0.281 & 0.272 & 0.272 & 0.274 & 0.253 & 0.241 & 0.233& 0.231 & 0.223 & 0.266\\
        LLaVA-NEXT-Video & 0.210 & 0.065 & 0.079 & 0.102 & 0.120 & 0.120 & 0.117 & 0.098& 0.097 &0.097 &0.114 \\
        MiniCPM-V 2.6 &0.418 &0.373 &0.352 &0.322 &0.308 &\textbf{0.302} &0.278 &0.284 &\textbf{0.272} &0.253 &0.318\\
        \rowcolor{mycolor}
        Ours(anls)  & \textbf{0.456} & \textbf{0.392} & \textbf{0.374} & \textbf{0.345} & \textbf{0.319} & 0.299 & \textbf{0.288} & \textbf{0.286}& 0.252& \textbf{0.259}  &\textbf{0.335}\\

        \hline
        \hline
        LongVA & 9.7 & 9.9 & 7.8 & 10.5 & 8.7 & 11.5 & 7.4& 8.0 & 6.7& 5.1 & 8.1  \\
        LongVILA-8B & 11.2 & 10.5 & 9.8 & 10.8 & 9.7 & 11.6 & 8.6 & 8.1& 7.9 & 6.2 &8.9  \\
        LLaVA-NEXT-Video & 5.0 & 2.0 & 0 & 0 & 0 & 0 & 0 & 0& 0 & 0& 1.0\\
        MiniCPM-V 2.6 & 16.2 & 14.1 & 13.9 & 11.7 & 10.7 & 10.8 & 9.9 & 9.8& 9.4 & 8.1  &11.6 \\
        \rowcolor{mycolor}
        Ours(acc)  & \textbf{23.8} & \textbf{21.2} & \textbf{18.9} & \textbf{16.3} & \textbf{13.4} & \textbf{12.4} & \textbf{12.2} & \textbf{12.1}& \textbf{11.2}& \textbf{11.0}  &\textbf{16.0}\\

        \hline
    \end{tabular}
    }
\end{table*}

\begin{table*}[h]
    \centering
    \caption{Comparative Analysis of Model Performance on Video Text Question Answering Across Diverse Scenarios \textbf{(Text Needle-in-Haystack Task)}. Accuracy is reported in percentage values (e.g., 2.4 denotes 2.4\%).}
    \label{tab:laozhen_cla}
    \resizebox{0.8\textwidth}{!}{
    \begin{tabular}{l|cccccccccc}
        \hline
        \textbf{\multirow{2}{*}{Models}}  & {\textbf{\multirow{2}{*}{Driving}}}& {\textbf{Egocen-}}& {\textbf{Enterta-}}&{\textbf{\multirow{2}{*}{Game}}}&{\textbf{Knowl-}}&{\textbf{Lifer-}}&{\textbf{\multirow{2}{*}{Sports}}}&{\textbf{\multirow{2}{*}{Talking}}}&\textbf{Video-}&{\textbf{\multirow{2}{*}{Avg}}}
        \\
         &  & {\textbf{tric}} & {\textbf{inment}} & {\textbf{}}& {\textbf{edge}}& {\textbf{ecord}}& {\textbf{}}& {\textbf{}}& {\textbf{news}} &\\
        \hline
        LongVA & 0.113 & 0.163 & 0.211 & 0.148 & 0.330 & 0.236 & 0.268 & 0.414 & 0.223 & 0.240  \\
        LongVILA & 0.186 & 0.094 & 0.342 & 0.256 & 0.320 & 0.250 & 0.280 & 0.301& 0.230 & 0.266 \\
        LLaVA-NEXT-Video & 0.079 & 0.102 & 0.102 & 0.137 & 0.201 & 0.104 & 0.101 & 0.159& 0.101 & 0.114 \\
        MiniCPM-V 2.6  & 0.180 & 0.275 & 0.444 & \textbf{0.425} & 0.387 & \textbf{0.251} & 0.302 & 0.439 & \textbf{0.332} & 0.318 \\
        \rowcolor{mycolor}
        Ours(anls)  & \textbf{0.229} & \textbf{0.301} & \textbf{0.460} & 0.372 & \textbf{0.420} & 0.231 & \textbf{0.324} & \textbf{0.471} & 0.321 &\textbf{0.335} \\
        \hline
        \hline
        LongVA & 2.4 & 2.9 & 3.9 & 1.1 & 10.1 & 3.8 & 11.5 & 23.6& 4.2 & 8.1  \\
        LongVILA-8B & 3.5 & 2.2 & 5.5 & 1.8 & 12.0 & 4.5 & 12.9 & 25.1& 5.0 & 8.9  \\
        LLaVA-NEXT-Video & 2.3 & 1.1 & 1.1 & 1.3 & 2.4 & 0.8 & 0.7 & 3.0& 0.90 & 1.0\\
        MiniCPM-V 2.6 & 3.8 & 6.1 & 21.7 & 23.2 & 13.1 & 5.7 & 12.6 & 20.0 & 10.3 & 11.6  \\
        \rowcolor{mycolor}
        Ours(acc)  & \textbf{7.9} & \textbf{8.8} & \textbf{22.5} & \textbf{25.8} & \textbf{18.3} & \textbf{7.5} & \textbf{15.4} & \textbf{28.5} & \textbf{17.0} &\textbf{16.0}\\

        \bottomrule
    \end{tabular}
    }
\end{table*}

\begin{table*}[!ht]
    \centering
    \caption{Performance on the \textbf{Text Time Grounding Task}. Top to bottom: model accuracy at 30s, 60s, and 120s fault tolerance. Average video length is 2306 seconds.}
    \label{tab:timeloc}
    \resizebox{0.8\textwidth}{!}{
    \begin{tabular}{l|ccccccccc}
        \hline
        \textbf{\multirow{2}{*}{Models}}  & {\textbf{\multirow{2}{*}{Driving}}}& {\textbf{Egocen-}}& {\textbf{Enterta-}}&{\textbf{\multirow{2}{*}{Game}}}&{\textbf{\multirow{2}{*}{Knowledge}}}&{\textbf{Life-re}}&{\textbf{\multirow{2}{*}{Sports}}}&{\textbf{\multirow{2}{*}{Talking}}}&{\textbf{Video-}}
        \\
         &  & {\textbf{tric}} & {\textbf{inment}} & {\textbf{}}& {\textbf{}}& {\textbf{cord}}& {\textbf{}}& {\textbf{}}& {\textbf{news}}\\
        \hline
        \multirow{3}{*}{LongVA}  
        &1.47 &1.93 &0.0 &2.56 &7.19 &2.23 &2.83 &0.0 & 1.14\\
        &3.47 &1.34 & 0.0 &9.31 &6.18 &2.73 &3.42 &0.0 &2.18\\
        &4.53 &2.71 &0.0 &10.51 &11.52 &4.33 &4.10 &6.29 &3.48\\\hline

        \multirow{3}{*}{LongVILA} 
        &0.95 &0.58 &0.08 &2.33 &3.52 &1.95 &1.64 &0.05 &0.77 \\
        &1.81 &0.57 &0.009  &8.55 &3.98 &2.78 &2.16 &0.0&1.18  \\
        &4.27 &1.45 &0.0 &9.63 &6.43 &2.83 &2.93 &0.71 &1.82 \\\hline

        \multirow{3}{*}{LLaVA-NEXT-Video}
        &0 &1.61 &0 &\textbf{6.15} &1.01 &1.62 &0.96 &0 & 0\\
        &0 &3.22 &0 &7.69 &5.05 &3.57 &1.92 &0 & 1.13\\
        &0 &5.30 &0 &15.38 &9.09 &7.14 &4.03 &0 & 2.27\\
        \hline

        
        \multirow{3}{*}{Qwen2-VL-7B}  
        &1.01 &1.14 &0.0 &3.09 &4.75 &1.64 &2.14 &0.0 &1.71 \\
        &2.32 &1.53 &0.0 &9.12 &5.82 &3.78 &2.87 &0.0 &1.75 \\
        &4.45 &3.47 &0.0 &11.12 &7.50 &5.42 &4.44 &1.14 &3.39 \\
        \hline

        \multirow{3}{*}{MiniCPM-V 2.6}  
        &1.31 &1.22 &0.0 &3.07 &5.05 &1.95 &2.11 &0.0 &1.42 \\
        &2.63 &1.63 &0.0 &9.23 &6.06 &3.57 &2.88 &0.0 &1.98 \\
        &5.26 &3.67 &0.0 &13.85 &9.10 &5.84 &4.23 &1.41 &3.69 \\
        \hline
        

        




        \rowcolor{mycolor}
        \multirow{3}{*}{\cellcolor{mycolor}Ours} 
        &\textbf{2.64} &\textbf{3.26} &\textbf{0.81}  &\textbf{6.15} &\textbf{11.22} &\textbf{6.81} &\textbf{3.46} &\textbf{1.92} &\textbf{5.39}\\
        \rowcolor{mycolor}
        Ours(acc)
        &\textbf{5.96} &\textbf{8.97} &\textbf{2.45} &\textbf{10.76} &\textbf{19.38} &\textbf{9.74} &\textbf{5.96} &\textbf{3.84} & \textbf{8.52}\\
        \rowcolor{mycolor}
        &\textbf{11.25} &\textbf{18.77} &\textbf{3.07} &\textbf{18.46} &\textbf{30.61} &\textbf{18.18} &\textbf{10.70} &\textbf{7.27} & \textbf{15.05}\\
        \hline
    \end{tabular} }
   
\end{table*}

\begin{table*}[!ht]
    \centering
    \caption{Performance comparison of different models on \textbf{Text Dynamics Caption} (points from 0 to 10)}
    \label{tab:textchange}
    \resizebox{0.8\textwidth}{!}{
    \begin{tabular}{l|ccc|ccc|c}
        \hline
        \textbf{\multirow{2}{*}{Models}} & \multicolumn{3}{c|}{\textbf{Sport}} & \multicolumn{3}{c|}{\textbf{Game}} & \textbf{\multirow{2}{*}{Avg}} \\ \cline{2-7}
        &basketball & ping-pong & swim & dota & lol& pubg  \\  
        \hline
        LongVA  &2.76 &2.74 &3.44 &2.36 &1.76 &2.84 &2.61 \\
        LongVILA &3.76 &{4.93} &2.61 &{3.70} &2.15 &{4.19} &{3.55}\\
        LLaVA-NeXT-Video &0.68 &2.62 &2.44 &2.19 &0.69 &1.63 &1.69\\
        Qwen2-VL-7B  &4.01 &4.03 &3.21 &3.73 &2.76 &2.55  &3.38\\
        MiniCPM-V 2.6   &3.81 &4.42 &4.47 &2.72 &3.26 &2.75  &3.44\\
        \rowcolor{mycolor}
        Ours  &\textbf{4.14} &\textbf{5.52} &\textbf{4.70} &\textbf{4.37} &\textbf{3.66} &\textbf{5.13}  &\textbf{4.59}\\

        \hline
    \end{tabular}
    }
\end{table*}

         
\subsection{Ablations}
\begin{table}[h]
\centering
\caption{Ablation experiments on time tasks were conducted for IT-RoPE and TimePrompt, wherein errors within 30 seconds, 60 seconds, and 120 seconds were considered correct.}
\label{tab:experrment_Effect_IT}
\begin{tabular}{l|ccc}
\hline
\textbf{Model}  
&30s &60s &120s 
\\ \hline
Ours &5.01 &8.24 &13.89 \\ 
w/o IT-RoPE &4.01 &6.64 &9.86 \\ 
w/o Time Prompt &0 &1.21 &2.13
\\ 
\hline
\end{tabular}
\end{table}

\begin{table}[h]
    \centering
    \caption{Ablation study of Non-uniform Position Interpolation on the Text-Neddle in Haystack task. Settings with duration greater than 14 minutes (sampling rate fps = 0.5) were selected to test its capability in handling long videos.}
    \label{tab:ab_nupi}
    \begin{tabular}{l|cccc}
        \hline
        \textbf{Models}  &{\textbf{14min}}&{\textbf{16min}}&{\textbf{18min}}&{\textbf{20min}} \\
        \hline
        Ours(anls)   & 0.278 & 0.286 & 0.252& 0.259   \\
        w/o NUPI & 0.260 & 0.253 & 0.231& 0.239 \\
        \hline
    \end{tabular}
\end{table}

\noindent \textbf{Effect of IT-RoPE.} This module enhances temporal localization performance by explicitly encoding frame sequential positions. Trained on VideoTime and EliteSet datasets, it is systematically evaluated on the BenchMark task. As shown in Table \ref{tab:experrment_Effect_IT}, the IT-RoPE-based model outperforms the original 1d-RoPE architecture. Further visualization analysis is provided in Appendix Section \ref{visattention}.

\noindent \textbf{Effect of Non-uniform Position Interpolation.}
Non-uniform Position Interpolation demonstrates superior performance when dealing with extremely long tokens. Therefore, we adopt this method when the number of sampled video frames exceeds 400. As shown in Table \ref{tab:ab_nupi}, the inclusion of Non-Uniform Position Interpolation (NUPI) leads to a higher ANLS metric on the Text-Neddle in Haystack task.

\noindent \textbf{Effect of Time Prompt.}
Table \ref{tab:experrment_Effect_IT} demonstrates that in the context of long video temporal understanding tasks, the time prompt is particularly crucial. Without the time prompt, the model almost entirely loses its temporal localization capability.

\noindent \textbf{Quantitative Results.}
Figure \ref{fig:case1},Figure \ref{fig:case2},Figure \ref{fig:case3}and Figure \ref{fig:case4} illustrate example outputs of the proposed model under three task settings. It can be observed that long videos undergo numerous scene transitions, and text within scenes is utilized in some video question answering tasks. The model demonstrates a strong capability in understanding text within long videos.

\section{Conclusion}
We present TextVidBench , the first benchmark that extends video-based scene text understanding evaluation from traditional clips of approximately 20 seconds to long-form videos exceeding three minutes in duration. To comprehensively assess scene text comprehension capabilities in such long videos, we design three challenging task settings:1)Text Needle-in-Haystack. 2) Text Temporal Grounding. 3)
Text Dynamics Captioning. These tasks are designed to simulate real-world applications, such as content retrieval and question answering in online educational video lectures, as well as dynamic commentary involving athlete actions and data evolution in sports events. In light of the unique characteristics of these tasks, we explore several key model improvements, including: temporal aggregation with rotary positional encoding, non-uniform positional encoding for enhanced long-token modeling, and precise alignment through absolute timestamp integration. Experimental results validate the effectiveness of these strategies. In summary, TextVidBench not only provides a systematic evaluation platform for text understanding in long video scenarios but also offers valuable methodological insights and references for future research in this field.

\section*{Limitations}

Due to the rapid evolution of architectures for multimodal large models, and considering that top-tier research institutions often have access to significantly stronger computational resources (e.g., massive GPU clusters), the improved model proposed in this paper may not fully match the performance of some recently emerged, heavily parameterized models trained on large-scale datasets that continue to appear on arXiv. Nevertheless, the methodological enhancements introduced in this work—particularly the design strategies tailored for long video-text understanding tasks—exhibit a certain level of generality and transferability. Future researchers can apply the improvements proposed in this paper to the latest base models and conduct iterative evaluations on the benchmark introduced herein, potentially achieving superior performance in subsequent studies.



\bibliography{custom}

\newpage       
\onecolumn

\appendix


\vspace*{1pt} 
\begin{center}
  {\LARGE\bfseries Appendix} \\ 
  \vspace{20pt} 
\end{center}

\section{Annotation Process}
\begin{figure}[ht]
\centering
\includegraphics[width=0.8\columnwidth]{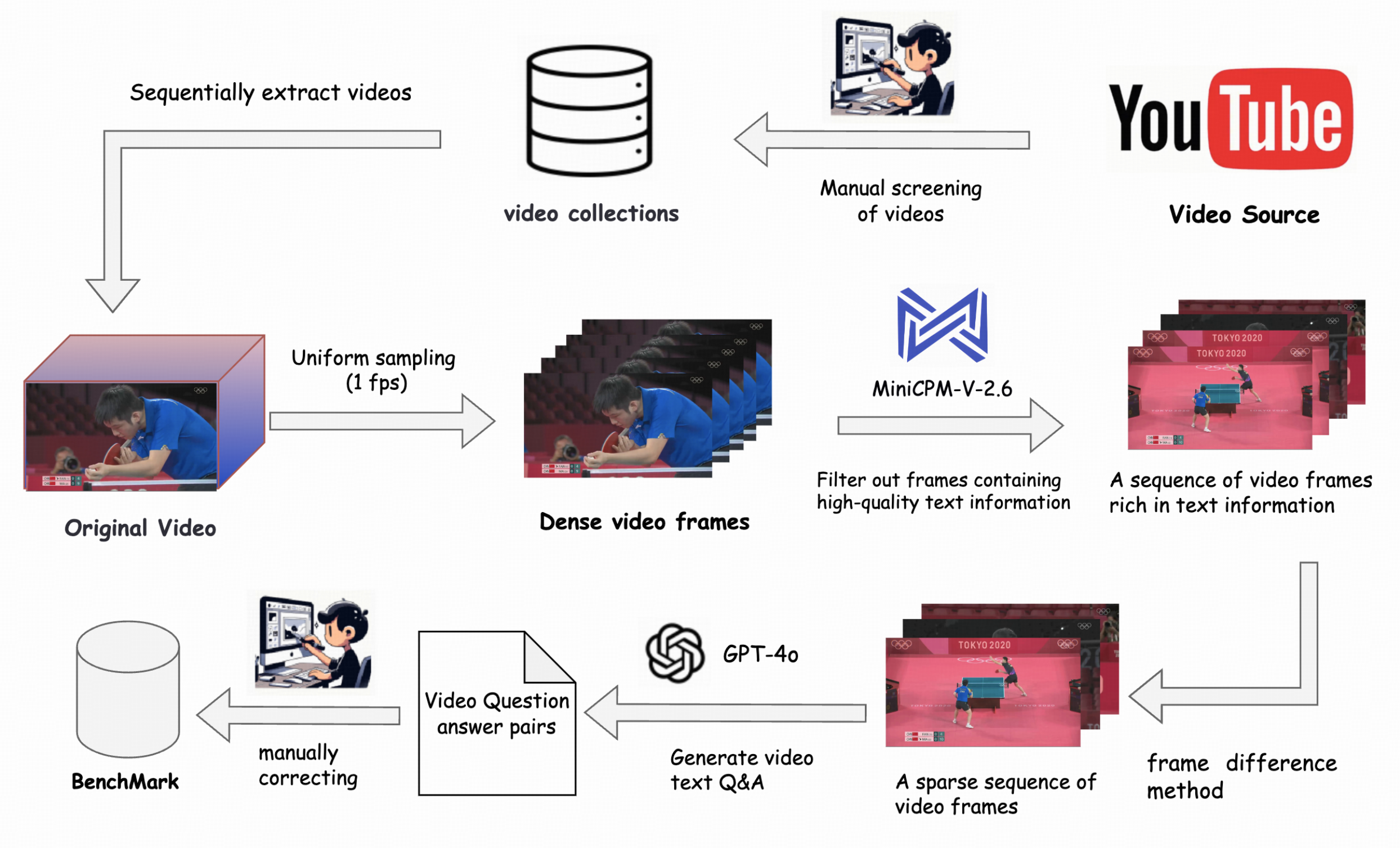} 
\caption{Semi-automatic Annotation Process.}
\label{fig:dataproduce}
\end{figure}

\begin{figure*}[h]
\centering
\includegraphics[width=\textwidth]{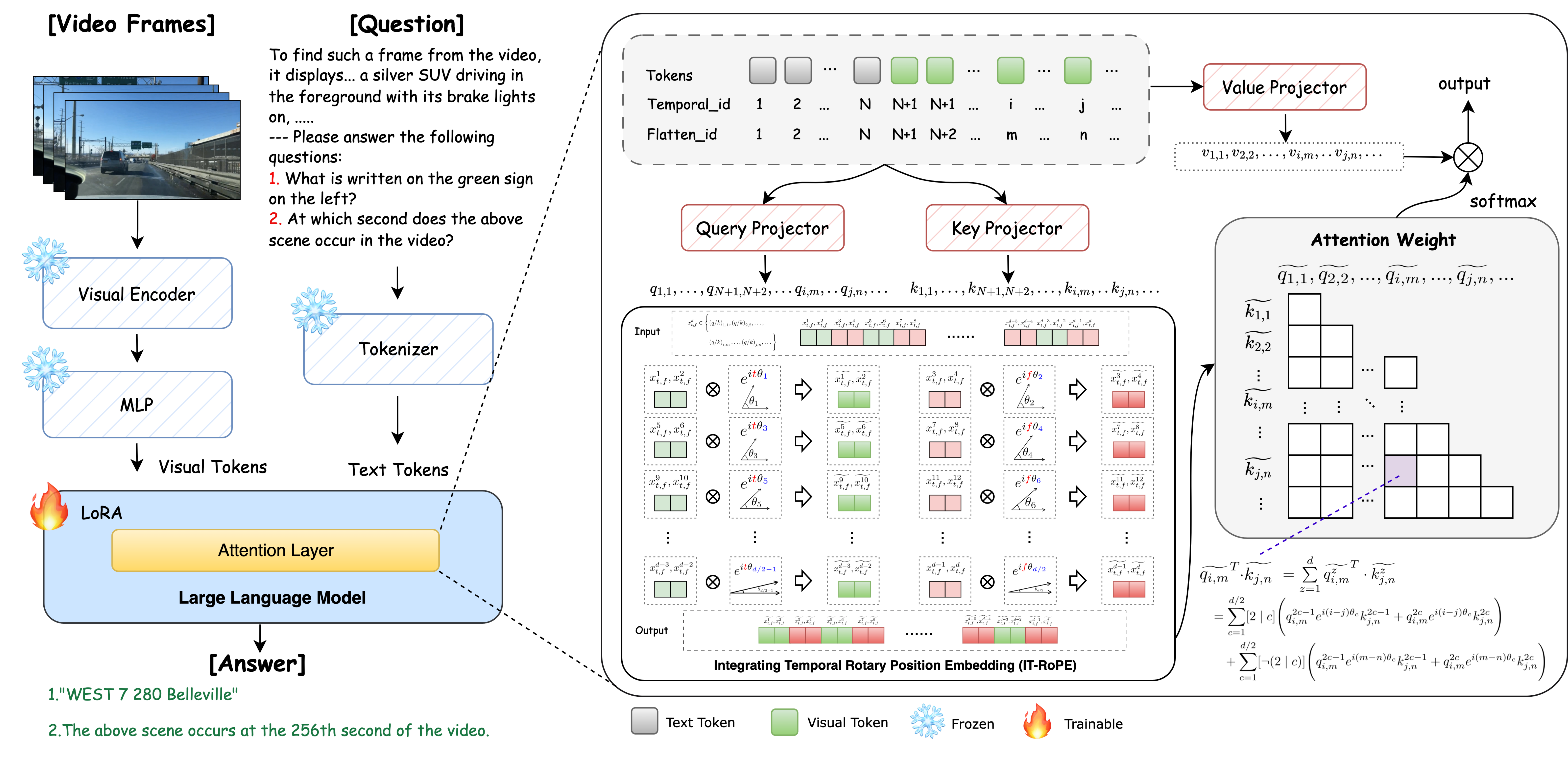}
\caption{The overall architecture of our model.}
\label{arch}
\end{figure*}
\section{Model Architecture}
\label{fulu:Model Architecture}
The model comprises three fundamental components: a visual encoder, a visual projector (MLP), and a large language model. The overall architecture is illustrated in Figure~\ref{arch}. The design of each module is as follows:

\begin{itemize}
    \item \textbf{Visual Encoder}: We employ the pre-trained visual encoder from MiniCPM-V2.6~\cite{yao2024minicpm}, which is based on the SigLIP SoViT-400m/14~\cite{zhai2023sigmoid} architecture.
    
    \item \textbf{Visual Projection Layer (MLP)}: This layer utilizes the pre-trained resample layer from MiniCPM-V2.6~\cite{yao2024minicpm}. It is designed to compress visual tokens, making them suitable for tasks with dense visual tokens such as video understanding, and to map visual features to textual features.
    
    \item \textbf{Large Language Model}: We adopt Qwen2-7B~\cite{qwen} as the foundational architecture for processing both video and text. In this study, we propose the IT-RoPE (For a detailed explanation, refer to Section~\ref{sec:itrope}) in the attention layer to extend the video sequence dimension, encoding the sequence of video frames into the tokens. Additionally, to better handle long videos, we have, for the first time, applied LongRoPE to perform hyperparameter search on the rotation matrix of Qwen2, thereby improving its capability to process long videos. (For a detailed explanation, refer to Section~\ref{sec:nonupi}).
\end{itemize}

\section{Statistics and Analysis}
\label{fulu:Statistics and Analysis}
\begin{figure*}[t]
  \centering
  \begin{subfigure}{0.3\linewidth}
    \includegraphics[width=\linewidth]{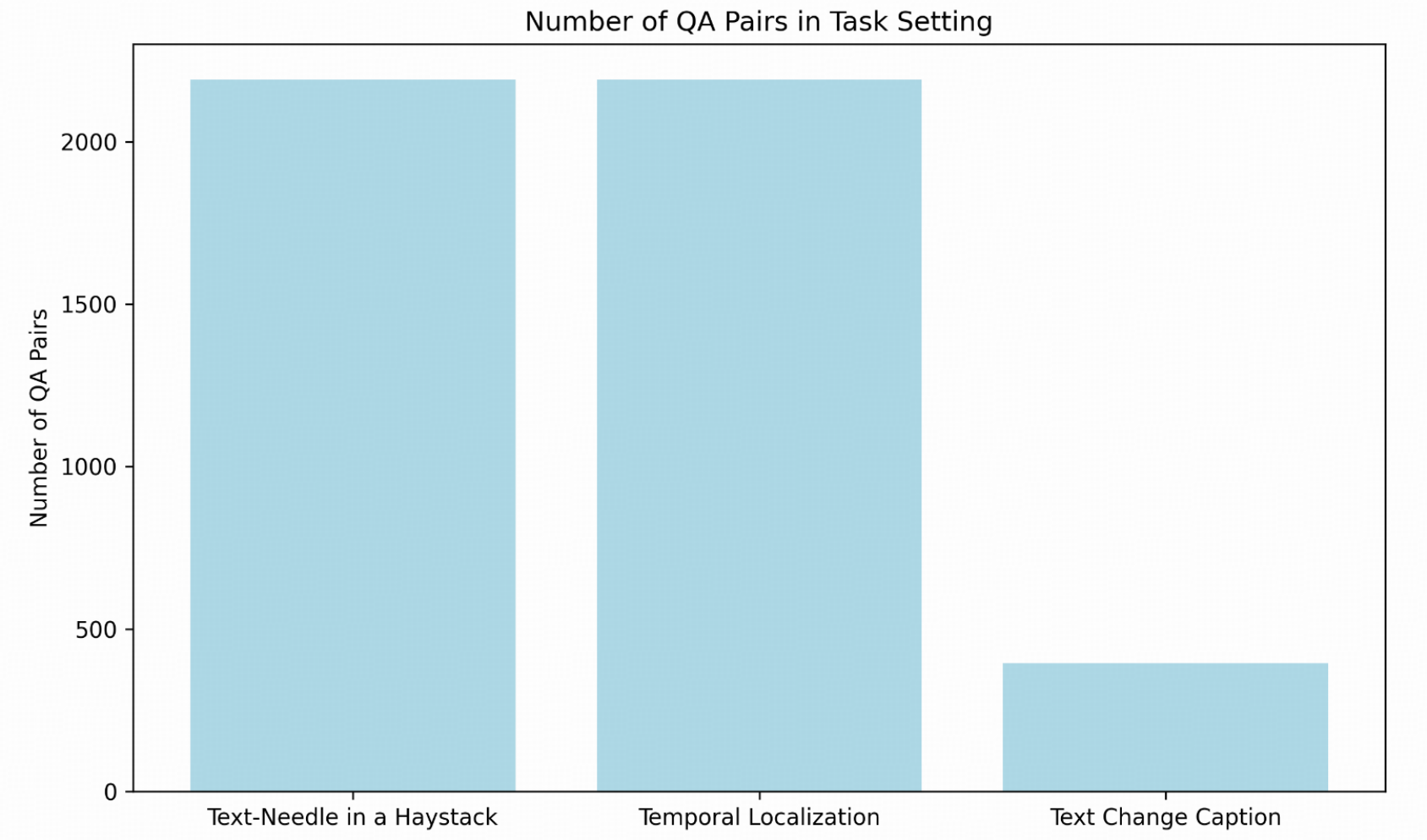}
    \caption{The number of question answer pairs for task settings.}
    \label{fig:short-a}
  \end{subfigure}
  \hfill
  \begin{subfigure}{0.3\linewidth}
  \centering
    \includegraphics[scale=0.2]{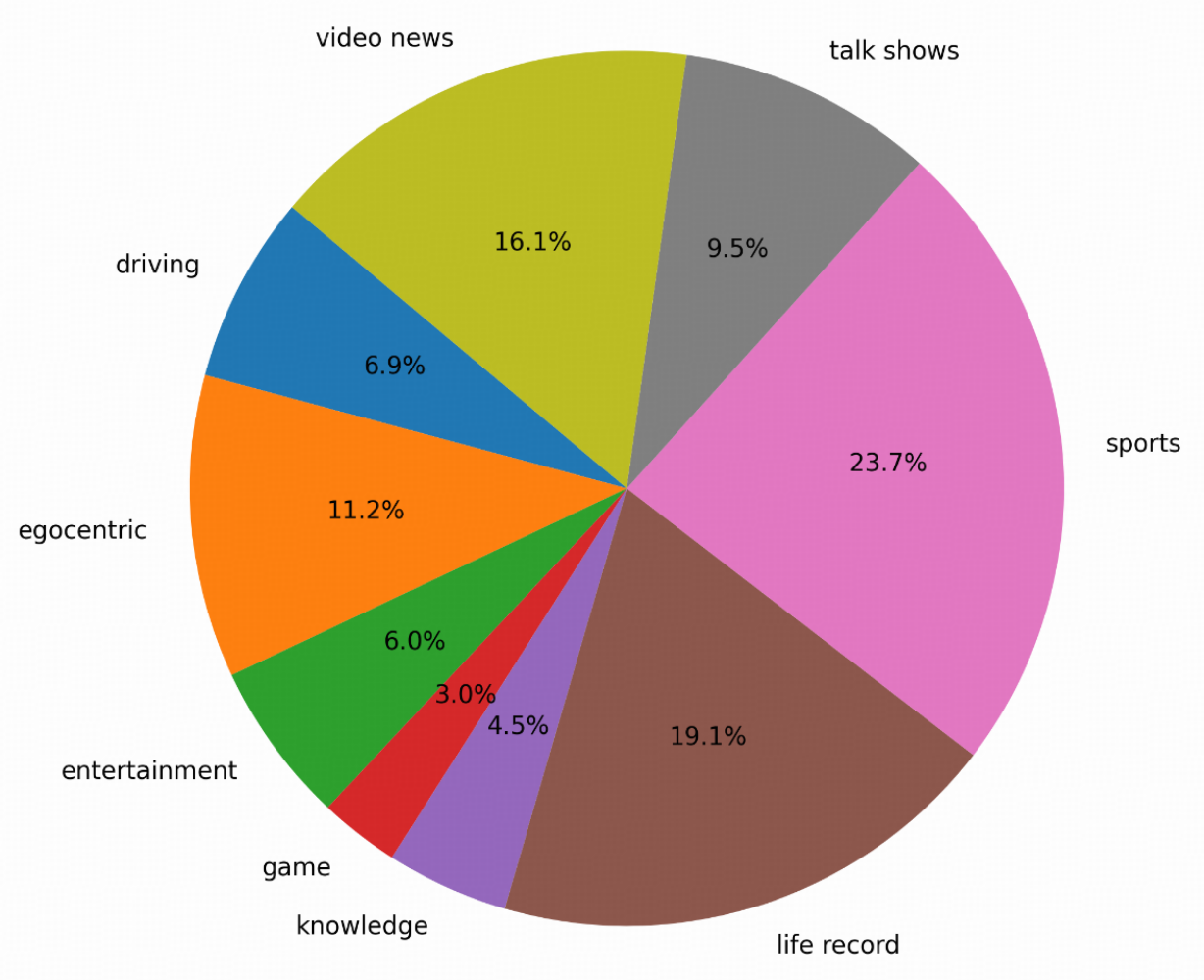}
    \caption{Proportion of 9 scenario types in Text-Needle in Haystack/Temporal Location}
    \label{fig:short-b}
  \end{subfigure}
  \hfill
  \begin{subfigure}{0.3\linewidth}
    \includegraphics[width=\linewidth]{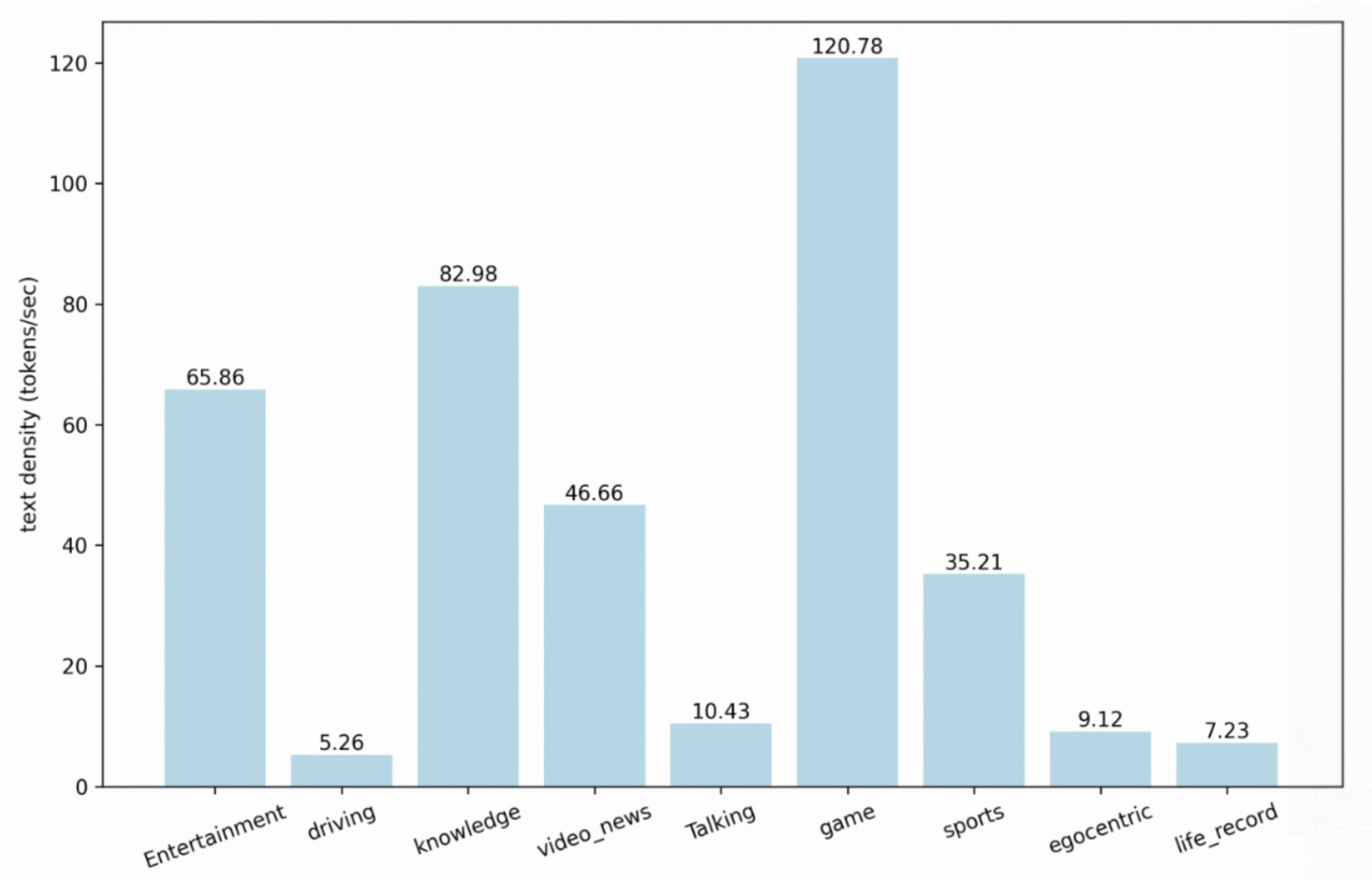}
    \caption{Text Density in Each Scene (Number of Text Tokens / Frame)}
    \label{fig:short-c}
  \end{subfigure}
  
  
  \begin{subfigure}{0.3\linewidth}
    \includegraphics[width=\linewidth]{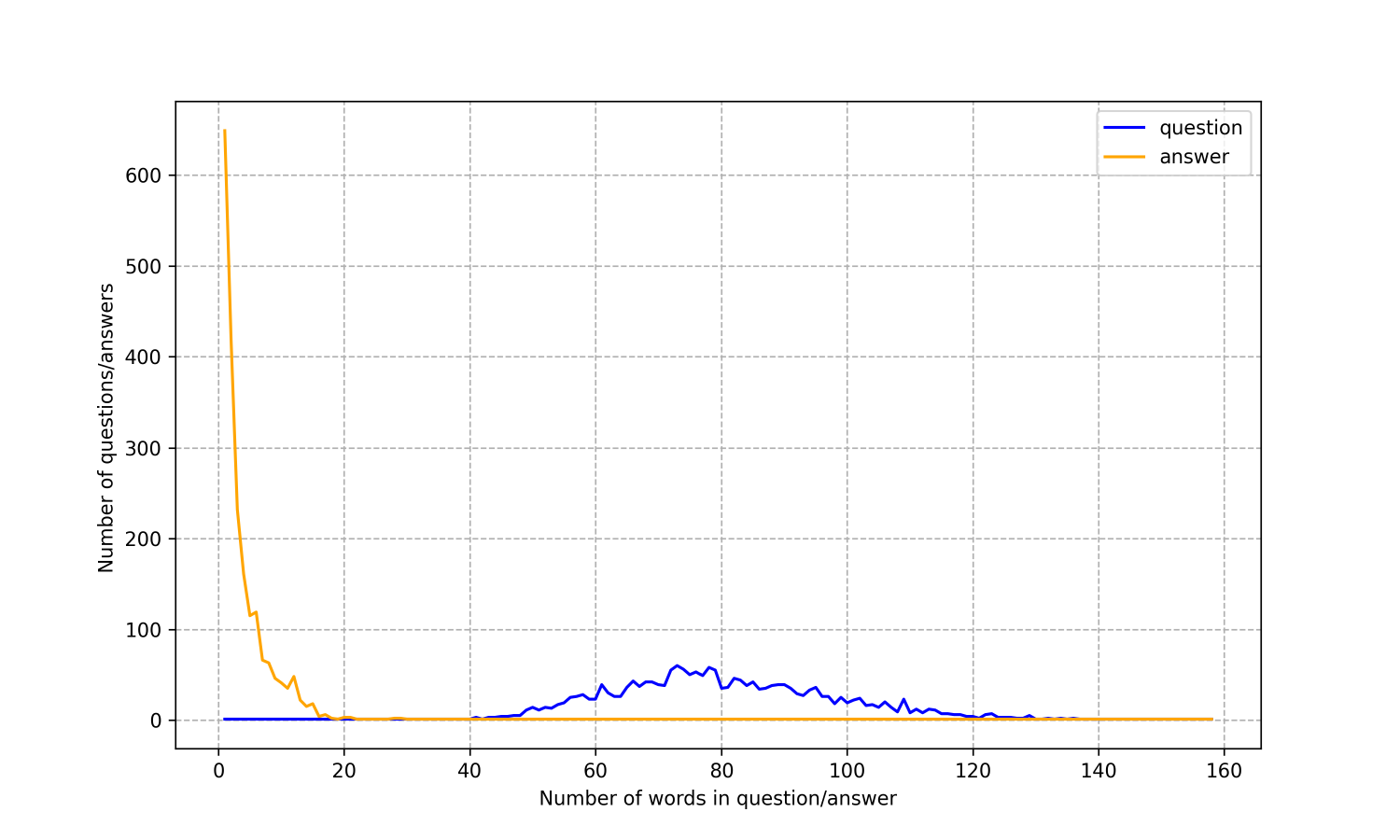}
    \caption{Statistical graph of the number of words in questions/answers.}
    \label{fig:short-d}
  \end{subfigure}
  \hfill
  \begin{subfigure}{0.23\linewidth}
  \centering
    \includegraphics[width=0.8\linewidth]{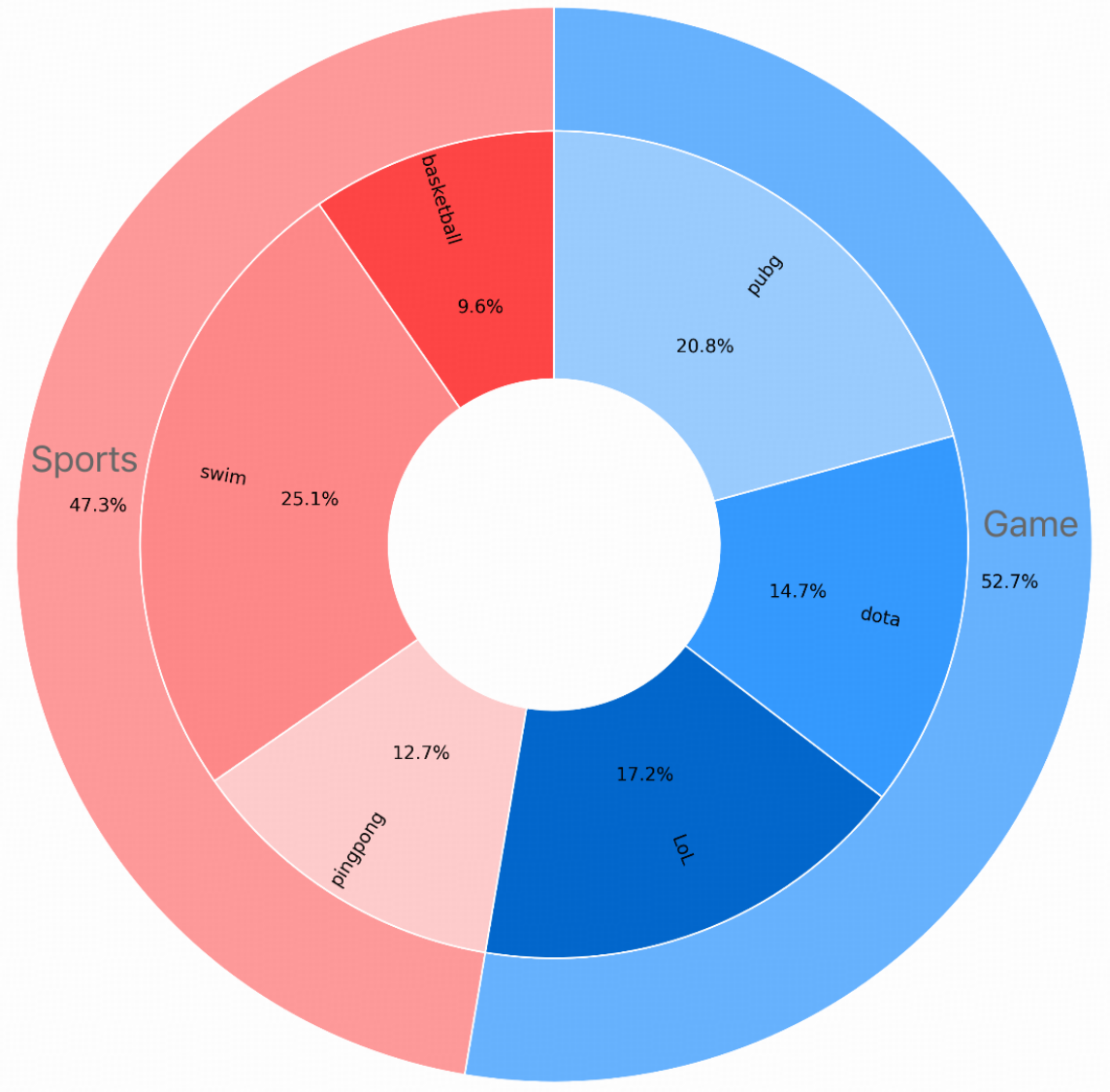}
    \caption{Distribution chart of scenarios covered by the Text Change task.}
    \label{fig:short-e}
  \end{subfigure}
  \hfill
  \begin{subfigure}{0.3\linewidth}
    \includegraphics[width=\linewidth]{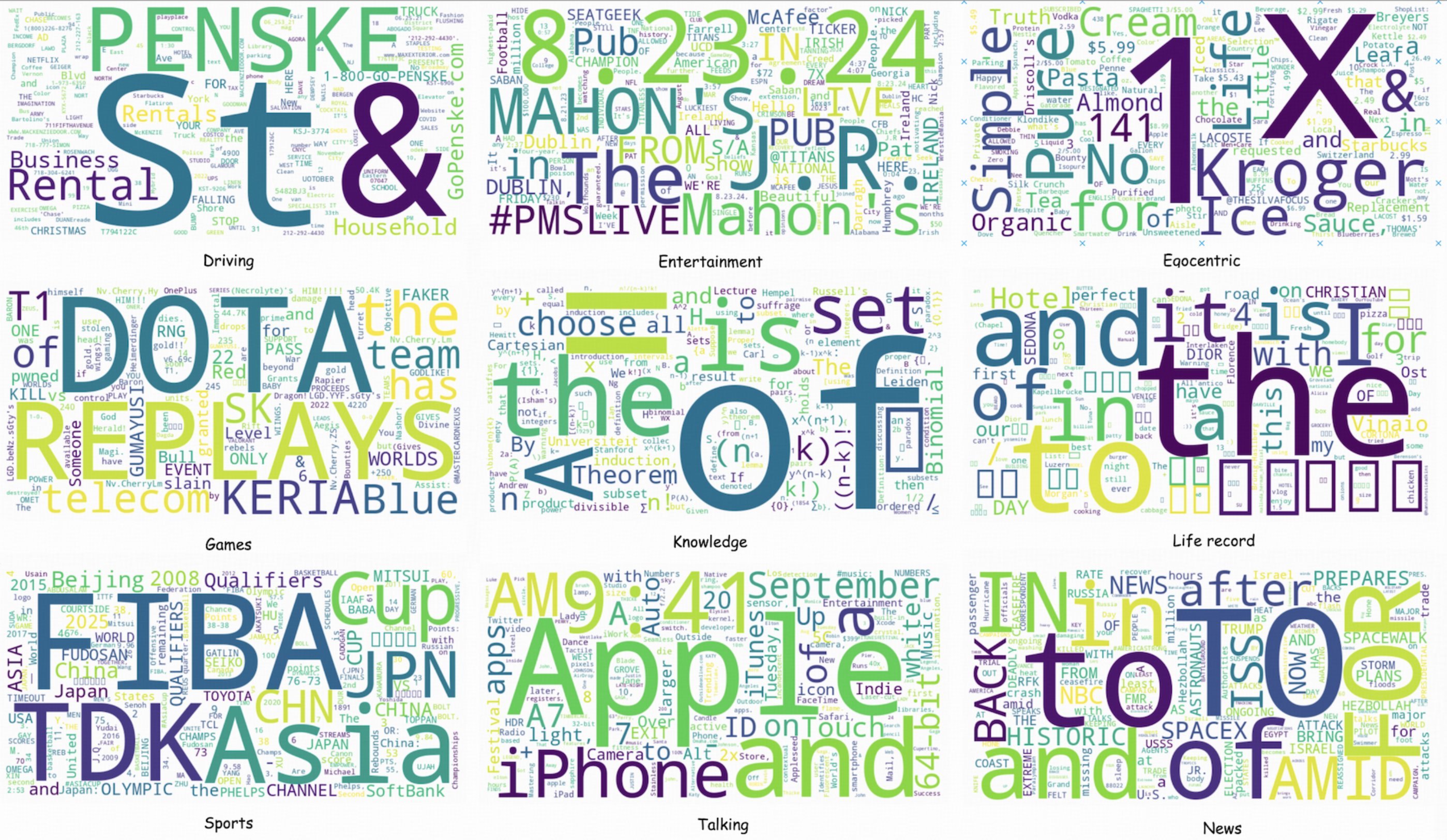}
    \caption{Word cloud diagram of answers in question-answer pairs across 9 types of scenarios.}
    \label{fig:short-f}
  \end{subfigure}
  
  \caption{Statistical data on benchmark.}
  \label{fig:short}
\end{figure*}
We presents a systematic analysis of the statistical characteristics of the constructed dataset. Figure \ref{fig:short-a} illustrates the data scale distribution across three task settings: the text-needle in haystack task comprises 2,190 question-answer pairs, corresponding to a total duration of approximately 20 hours (with an average video length of 2,306 seconds); the temporal location task extends the original task by reformulating the questioning and answering methods of text-needle (transforming text retrieval questions into temporal localization questions), while preserving the original video distribution (the proportion of 9 scene categories is shown in Figure \ref{fig:short-b}) and duration characteristics; the text change caption task, addressing the need for describing semantic dynamic changes, carefully selects 600 question-answer pairs covering sports (basketball/ping-pong/swimming) and esports (LOL/PUBG/DOTA) scenarios, with detailed subclass distribution shown in Figure \ref{fig:short-e}. Compared to existing video understanding benchmark datasets, this work achieves significant breakthroughs in multiple dimensions: the average video duration is increased by approximately 100 times compared to previous evaluation benchmarks \cite{2022Watching,tom2023readinglanestextvideoqa,zhao2022towards}, and it innovatively integrates fine-grained temporal localization and dynamic semantic capture tasks.

To further reveal the data characteristics, Figure \ref{fig:short-c} presents an analysis of text information density (tokens/s) across nine scene categories, showing that gaming and knowledge-based scenes have significantly higher text embedding densities (with means of 5.8/s and 4.6/s, respectively), while driving scenes have the lowest density (1.2/s).

In terms of question complexity, statistics in Figure \ref{fig:short-d} indicate that the average question length in this study reaches 78.6 words, which is 7 times longer than previous benchmarks \cite{2022Watching,tom2023readinglanestextvideoqa,zhao2022towards}, posing higher demands on the model's semantic understanding capabilities. Finally, through the visualization of answer word clouds across nine scene categories (Figure \ref{fig:short-f}), the distribution of scene-specific vocabulary can be intuitively observed.

\section{Details of IT-RoPE}
\label{Details of IT-RoPE}
Let the input consist of a text sequence followed by a sequence of images. The text is tokenized into \( T = \{t_1, t_2, \dots, t_N\} \), where \( N \) is the number of text tokens. Each image is tokenized into \( K \) tokens, such that the \( j \)-th image is represented as \( I_j = \{i_{j,1}, i_{j,2}, \dots, i_{j,K}\} \). For \( M \) images, the total number of image tokens is \( M \times K \). In the original 1-D position encoding scheme, the text and image tokens are flattened into a single sequence. The position encoding for the entire sequence is assigned as eq~\ref{eq_flattenid}:
\begin{equation}\label{eq_flattenid}
P_{\text{Flatten\_id}} = \{1, 2, \dots, N, N+1, N+2, \dots, N + M \times K\},
\end{equation}
where the text tokens occupy positions \( 1 \) to \( N \), and the image tokens occupy positions \( N+1 \) to \( N + M \times K \).

In our proposed approach, we modify the position encoding such that all tokens belonging to the same image share the same position encoding. Specifically:
\begin{itemize}
    \item The text tokens retain their original position encodings \( 1 \) to \( N \).
    \item For the image tokens, the \( K \) tokens of the \( j \)-th image are assigned the same position encoding \( N + j \).
\end{itemize}

Thus, the modified position encoding \( P_{\text{Temporal\_id}} \) is defined as:
\begin{equation}\label{eq_temporalid}
\begin{split}
P_{\text{Temporal\_id}} = \{ &1, 2, \dots, N, \underbrace{N+1, N+1, \dots, N+1}_{K \text{ times}}, \\
&\underbrace{N+2, N+2, \dots, N+2}_{K \text{ times}}, \dots, \\
&\underbrace{N+M, N+M, \dots, N+M}_{K \text{ times}}\}.
\end{split}
\end{equation}

The  position encoding \( P_{\text{Temporal\_id}}(i) \) for the \( i \)-th token can be formally expressed as:
\begin{equation}\label{tmporal2}
P_{\text{Temporal\_id}}(i) = 
\begin{cases} 
i & \text{if } 1 \leq i \leq N, \\
N + \left\lfloor \frac{i - N - 1}{K} \right\rfloor + 1 & \text{if } i > N,
\end{cases}
\end{equation}
where \( \lfloor \cdot \rfloor \) denotes the floor function. For text tokens (\( 1 \leq i \leq N \)), the position encoding is simply their index \( i \). For image tokens (\( i > N \)), the position encoding is determined by the image index \( j \), calculated as \( j = \left\lfloor \frac{i - N - 1}{K} \right\rfloor + 1 \). This ensures that all \( K \) tokens of the \( j \)-th image are assigned the same position encoding \( N + j \).

Specifically, for the feature vector \( X \) of a given input token, the positional indices are computed via equation \eqref{eq_flattenid} \eqref{eq_temporalid}, yielding \( P_{\text{Temporal\_id}} = t \) and \( P_{\text{Flatten\_id}} = f \). This can be represented as \( X_{(t,f)} \in \mathbb{R}^d \), where \( d \) denotes the feature dimension, \( t \) is the index along the temporal position, and \( f \) is the index along the flattened position, its rotational position encoding is as shown in Eq \eqref{eq4}.

\begin{equation}\label{eq4}
\tilde{X}_{t,f} = \text{IT-RoPE}(X_{t,f}) = R_{(t,f)}^{(d \times d)} X_{t,f}  .
\end{equation}

\noindent The matrix \( R_{(t,f)}^{(d \times d)} \) is a rotation matrix \( d \times d \), whose specific form is shown in Equation~\eqref{eq5}.

{\tiny 
\setlength{\arraycolsep}{2pt} 
\begin{equation}\label{eq5} 
R^{d \times d}_{t,f} =
\begin{pmatrix}
\cos t\theta_1 & -\sin t\theta_1 & 0 & 0 & \cdots & 0 & 0 \\
\sin t\theta_1 & \cos t\theta_1 & 0 & 0 & \cdots & 0 & 0 \\
0 & 0 & \cos f\theta_2 & -\sin f\theta_2 & \cdots & 0 & 0 \\
0 & 0 & \sin f\theta_2 & \cos f\theta_2 & \cdots & 0 & 0 \\
\vdots & \vdots & \vdots & \vdots & \ddots & \vdots & \vdots \\
0 & 0 & 0 & 0 & \cdots & \cos f\theta_{d/2} & -\sin f\theta_{d/2} \\
0 & 0 & 0 & 0 & \cdots & \sin f\theta_{d/2} & \cos f\theta_{d/2}
\end{pmatrix}
\end{equation}
}

\noindent This matrix groups features in pairs, dividing them into \( d/2 \) subspaces to form a block diagonal matrix for two-dimensional rotation. The sequence of subspaces is \( \{1, 2, \ldots, d/2\} \). For \( i \in \{1, 2, \ldots, d/2\} \), the rotation angle \( \theta_i=10000^{-2(i-1)/d} \) .For even values of \( i \), the corresponding rotation angle is given by \( k\theta_i \). When \( i \) is odd, the corresponding rotation angle is given by \( t\theta_i \).

In the process of attention computation within the attention heads of the Transformer model, for the feature vectors $q_{i,m}$ and $k_{j,n}$ derived from the query and key, respectively (where $i,j$ denote the indices on position ids (temporal) and $m,n$ denote the indices on position ids (flatten)), the attention scores before the softmax are defined using the dot product as Eq \eqref{eq6}

\begin{equation}\label{eq6}
\begin{split}
\tilde{q}_{i,m} \cdot \tilde{k}_{i,n} 
&= \left( R_{i,m} q_{i,m} \right)^\top \left( R_{j,n} k_{j,n} \right) \\
&= q_{i,m}^\top R_{(i-j,m-n)} k_{j,n}
\end{split}
\end{equation}

After simplification, it can be seen that only a rotation matrix \( R_{(i-j,m-n)} \) remains, then substitute $t = i - j$ and $k = m - n$ in Eq \eqref{eq5}. It can be observed that the distance between two positions depends solely on their relative positions and simultaneously integrates positional information from two dimensions.

Additionally, this type of position indexing automatically degenerates into one dimension when facing text tokens, which can greatly preserve the original text capabilities. For example, given two text tokens from positions \((a,a)\) and \((b,b)\), their \( R_{(a-b,a-b)} \) is equivalent to \( R_{(a-b)} \) in RoPE~\cite{su2024roformer}.

\section{Details of Non-uniform Position Interpolation}
\label{Details of Non-uniform Position Interpolation}
The long text processing capability of IT-RoPE can still be further enhanced. Inspired by the research of LongRoPE~\cite{longrope}, we have introduced the Non-uniform Position Interpolation technology, which is the first of its kind in large-scale language models (LLM) based on qwen~\cite{qwen}.

Specifically, the Rotation matrix\eqref{eq5} of IT-RoPE can be simplified to a sequence of cosine and sine functions, as shown below:

\begin{equation}\label{eq7}
\begin{split}
\bigl[ &\cos(t\theta_1), \sin(t\theta_1), \cos(f\theta_2), \ldots, \\
       &\cos(f\theta_{d/2}), \sin(f\theta_{d/2}) \bigr]
\end{split}
\end{equation}

By extending the original length $L$ to $L'$, we perform downsampling by introducing a hyperparameter $\lambda$ before each rotation frequency $\theta_i$, allowing the extended length to fit the length that the model has been trained on. This process can be described by the eq\eqref{eq8}. 
\begin{equation}\label{eq8}
\begin{split}
\biggl[ &\cos(\lambda_1 t\theta_1),\, \sin(\lambda_1 t\theta_1),\, \cos(\lambda_2 k\theta_2),\, \ldots, \\
        &\cos(\lambda_{d/2} k\theta_{d/2}),\, \sin(\lambda_{d/2} k\theta_{d/2}) \biggr]
\end{split}
\end{equation}

In the traditional linear positional interpolation (PI) method, all $\lambda_i$ are fixed as $L/L'$. However, as LongRoPE~\cite{longrope} points out, $\lambda_i$ can actually be non-uniform and have a large adjustment space. Therefore, we adopted the evolution search method~\cite{guo2020single} mentioned in LongRoPE to optimize $\lambda_i$ in the IT-RoPE. By this method, we found a set of optimized hyperparameters $\lambda_i$, significantly improving the model's performance.

\section{Video Instruction Data Generation and Tuning}
\label{instructiongeneration}

In this section, we will expound on the methodology to generate the video instruction dataset as well as the approach to tuning utilized in this context.

\vspace{1mm}


\subsection{Instruction Generation}

This section elaborates on the methodology and content of the proposed instruction dataset. The dataset is sourced from diverse origins, including video data crawled from the web combined with automated generation pipelines, as well as extensive modifications to existing public datasets. This results in a diverse set of scenarios covering multiple domains (e.g., knowledge, sports, news, etc.). The dataset encompasses various task types, such as text-based visual question answering, temporal localization, and video captioning. We have successfully constructed a high-quality dataset comprising 50,000 video-instruction pairs. This large-scale, multi-task dataset provides a solid foundation for model fine-tuning, enabling efficient understanding and reasoning of video textual content through deep integration of text and temporal cues. Specifically, as follows:

\begin{itemize}
    \item \textbf{VideoTextCap}: This dataset primarily focuses on caption generation for video textual information. We first crawled YouTube to download videos from nine categories rich in textual content, including driving, egocentric, entertainment, game, knowledge, life record, news, sports, and talking. These videos were randomly segmented into 1-5 minute clips. Suitable prompts were constructed to invoke Qwen-VL-72B~\cite{qwen2vl}, assisted by an OCR model, to generate captions emphasizing textual information in the video clips. The generated captions were then fed back into Qwen-VL-72B to simulate user-assistant dialogues, generating additional natural Q\&A pairs about the textual content in the videos. Finally, the data was manually verified to produce the final dataset. This dataset includes 40 hours of video, 800 video clips, and 800 Q\&A pairs, with an average clip duration of 3 minutes.

    \item \textbf{VideoTextQA}: This dataset focuses on natural Q\&A about video textual content, where correct answers require understanding the textual information in the videos. We first collected public datasets on video text comprehension, including M4-ViteVQA ~\cite{zhao2022towards}, RoadTextVQA ~\cite{tom2023readinglanestextvideoqa}, and NewsVideoQA ~\cite{2022Watching}. These datasets cover multiple scenarios, and we randomly selected six scenarios to concatenate into longer videos with more complex content, which helps train more robust models. This dataset contains 35,000 Q\&A pairs, with an average clip duration of 30 seconds.

    \item \textbf{VideoTime}: This dataset emphasizes temporal localization in video frames. Supplementing training with this dataset helps models better perform tasks related to video text Q\&A, such as describing temporal changes in textual information or retrieving specific scenes with textual content. The videos in this dataset are sourced from the video retrieval dataset DiDeMo~\cite{didemo}. The recent method VERIFIED~\cite{chen2024verified} optimized its captions for retrieval and provided quality scores for each caption. We collected captions with VERIFIED scores above 3.5 and constructed Q\&A pairs focused on temporal localization. This dataset contains 3,000 Q\&A pairs.

    \item \textbf{EliteSet}: The aforementioned dataset comprehensively enhances the model's ability to utilize textual information in videos and its temporal localization capabilities. However, its instruction-answering format significantly differs from the proposed benchmark. Therefore, based on the video data collected in VideoTextCap, we adopted the same data production process as the benchmark to create a small batch of question-answer pairs, approximately 800 in total. This portion of the data was fine-tuned separately after the initial fine-tuning of the first three sets of data.
\end{itemize}

\begin{table}[ht]
\centering 
\caption{Video Instruction Question Answer Format.}
\label{tab:input_sequence}
\begin{tabular}{ll} 
\hline
\textbf{User}: Instruction$_1$ + [Frames] & \textbf{Assistant}: A$_1$  \\
\textbf{User}: Instruction$_2$  & \textbf{Assistant}: A$_2$  \\
\ldots & \ldots\\
\textbf{User}: Instruction$_k$  & \textbf{Assistant}: A$_k$  \\
\hline
\end{tabular}
\label{table1}
\end{table}

\subsection{Instruction Tuning}
As shown in the Table\ref{table1}, we extract frames from the videos in the training data and combine them with several dialogue sequences to form a multi-modal instruction format. If there are multiple rounds of dialogue, the User input in the first round includes the video frame sequence and the user's command, with the corresponding model response denoted as A$_1$. In subsequent rounds, the User input contains only the text data of the user's command. We employ instruction-tuning of the Large Language Model (LLM) on the prediction tokens, leveraging its original auto-regressive training objective.The model's inference process is depicted in Equation \ref{eq2}, while the optimization formula is presented in Equation \ref{eq3}.

\begin{equation}\label{eq2}
p(A|I,F) = \prod_{k=1}^{Lenth} P_{\phi}(A_k |F,I_{,<k}, A_{,<k})  .
\end{equation}
\noindent Here, $I_{,<k}$ and $A_{,<j}$ denote the instruction and answer tokens, respectively, $F$ denote video frames and the current $k_{th}$ answer $A_k$. is predicted by previous inputs $I_{,<k}$ and $F$.

\begin{equation}\label{eq3}
\max_{\phi}\sum_{(F,I,A)\in Z}\sum_{k=1}^{|A|}\log(P_{\phi}(A_k\mid F,I, A_{<k})) .
\end{equation}

\noindent Here, the symbol $\phi$ represents the trainable parameters of our model,  $Z={(I_i ,A_i)}_{i=1,2,...,N}$ denotes a set comprised of pairs of instructions I and the corresponding answers A.

\section{Implementation Details}
\label{Implementationdetailes}
We employed the proposed model architecture and fine-tuned it on our newly introduced dataset comprising 50,000 video-instruction pairs. Specifically, we performed LoRA~\cite{lora} fine-tuning exclusively on the Large Language Model (LLM) layer, while keeping the remaining parts of the model architecture frozen. During training, we configured the following hyperparameters: a learning rate of $1 \times 10^{-6}$, weight decay of $0.1$, warmup ratio of $0.01$, and a cosine learning rate scheduler~\cite{cosine}. The fine-tuning process spanned 3 epochs and was executed on 4 NVIDIA A100 80GB GPUs, totaling approximately 30 hours of computation time.

\section{Validation on Proposed Benchmark}
\label{Validation on Proposed Benchmark}
\noindent{\textbf{1) Text-Needle in a Haystack:}}
To thoroughly investigate the model's text comprehension capabilities across varying video durations and diverse scenarios, we designed two sets of experimental tables. As shown in Table \ref{tab:laozhen_time}, we divided the videos into intervals of 2 minutes, 4 minutes, 6 minutes, up to 20 minutes, sampling is performed at 0.5 fps.. The video segments were extracted using a random sampling method that ensures each segment contains the timestamp corresponding to the current question-answer pair. This design allows each question-answer pair to be fully validated under different duration settings, thereby ensuring the scientific rigor and reliability of the experimental results. Table \ref{tab:laozhen_cla} further demonstrates the model's performance across 9 distinct scenarios, where the results for each scenario are averaged over the outcomes from 2-minute, 4-minute, up to 20-minute intervals. From the two sets of experimental results, it is evident that our proposed model significantly outperforms the current state-of-the-art models in this task on both the ANLS (Average Normalized Levenshtein Similarity) and ACC (Accuracy) metrics, thoroughly validating the effectiveness of the model.

\noindent{\textbf{2) Text Temporal Grounding:}} As shown in Table \ref{tab:timeloc}, we present the experimental results for the temporal localization task. In the experimental dataset, the average duration of each video is 2306 seconds. Considering the relatively long duration, a certain margin of error is allowed, and three tolerance levels are set: 30s, 60s, and 120s. We adopt a uniform sampling strategy, extracting 64 frames from each video for experimentation. The experimental results indicate that without using the time prompt method proposed in Section \ref{sec:timeprompt}, the localization accuracy of all compared models is nearly zero. Therefore, we incorporate the time prompt mechanism into all compared models. From the experimental results, it is evident that for long videos exceeding 3 minutes, most current video understanding models lack basic temporal localization capabilities. In contrast, our model, through improvements to RoPE (Rotary Position Embedding), effective application of time prompt, and the construction and training of a specific instruction set, achieves significant performance improvements in the task of temporal localization for long videos, demonstrating a certain level of temporal localization ability.

\noindent{\textbf{3) Text Dynamics Captioning:}}
As shown in Table \ref{tab:textchange}, we present the experimental results on the text change description task. All selected models are state-of-the-art multimodal large language models with approximately 7B parameters. The evaluation scores range from 0 to 10. The results indicate that the dynamic text description task remains a significant challenge for current models, as none of them achieve an average score above 5. This task comprehensively evaluates the model's ability to understand textual content in videos, requiring accurate text recognition, temporal localization, and long video sequence modeling. This evaluation setup provides a clear direction for future improvements in multimodal large models' capability to effectively utilize textual information in video understanding.

\section{Visualization of the attention scores}
\label{visattention}
\begin{figure}[h]
\centering
\includegraphics[width=0.99\columnwidth]{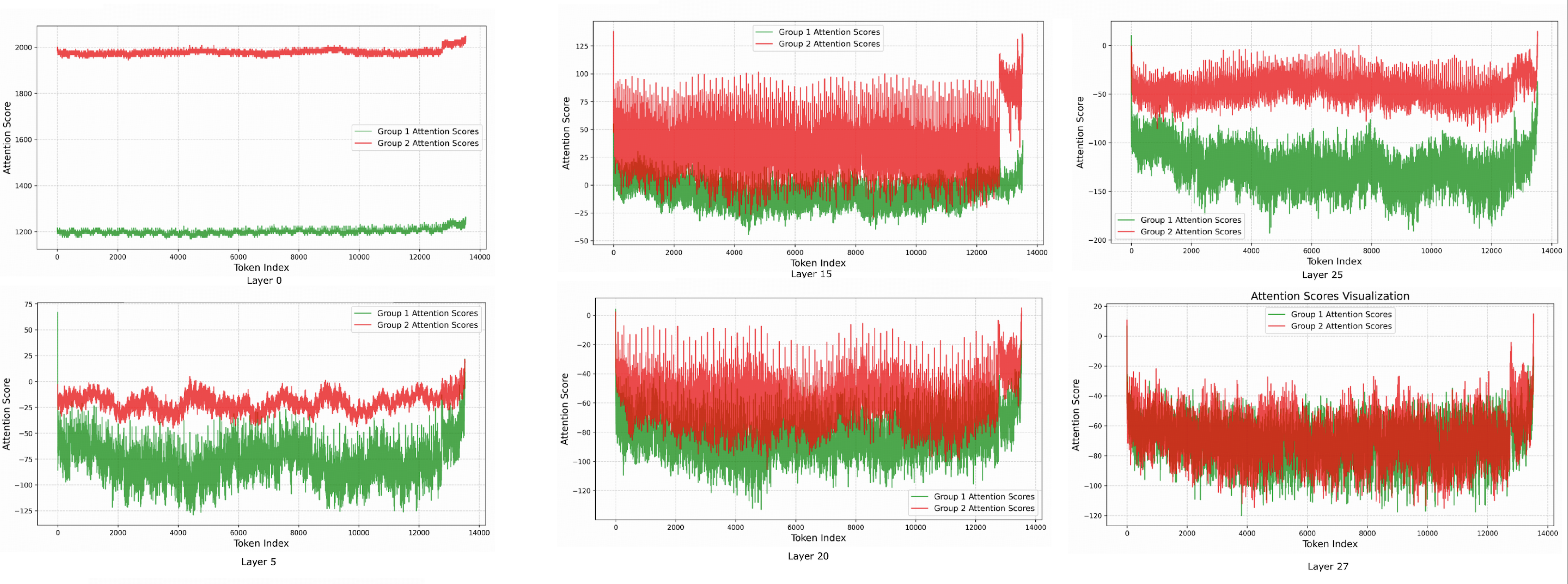} 
\caption{Visualization of attention scores across different transformer layers for various channel groups, where group 1 corresponds to the feature channel group associated with Flatten\_id during IT-RoPE encoding, and group 2 corresponds to the feature channel group associated with Temporal\_id.}
\label{fig:vis_attetntion}
\end{figure}

Figure \ref{fig:vis_attetntion}shows the attention scores between the text tokens output by the model and the visual tokens input to the model across different layers. It can be observed that, when performing the video temporal localization task, the feature channels corresponding to Temporal\_id play a more significant role across various layers of the Transformer.

\section{Quantitative Results}
\label{model_output}
Figure \ref{fig:case1},Figure \ref{fig:case2},Figure \ref{fig:case3}and Figure \ref{fig:case4} demonstrates the model's performance across various tasks, illustrating that the model has developed a certain capability in understanding text within long videos while also retaining its general video comprehension ability.

\begin{figure}[h]
\centering
\includegraphics[width=0.99\columnwidth]{fig/fulu/case1.pdf} 
\caption{Visualization of model output.}
\label{fig:case1}
\end{figure}

\begin{figure}[h]
\centering
\includegraphics[width=0.99\columnwidth]{fig/fulu/case2.pdf} 
\caption{Visualization of model output.}
\label{fig:case2}
\end{figure}

\begin{figure}[h]
\centering
\includegraphics[width=0.99\columnwidth]{fig/fulu/case3.pdf} 
\caption{Visualization of model output.}
\label{fig:case3}
\end{figure}

\begin{figure}[h]
\centering
\includegraphics[width=0.99\columnwidth]{fig/fulu/case4.pdf} 
\caption{Visualization of model output.}
\label{fig:case4}
\end{figure}

\end{document}